\documentclass[11pt]{elsarticle}
\usepackage{algpseudocode}
\usepackage{multirow}
\usepackage{booktabs}
\usepackage{array}
\usepackage{hyperref}
\usepackage{latexsym}
\usepackage{graphicx}
\usepackage{epsfig}
\usepackage{url}
\usepackage{amsmath,amssymb}
\numberwithin{equation}{section}
\usepackage{tabularx}
\usepackage{mathrsfs}
\usepackage{algorithm}
\usepackage{soul}
\usepackage{color}
\usepackage{subcaption}
\usepackage{enumerate}
\usepackage{booktabs}
\usepackage{svg}
\usepackage{cleveref}
\usepackage{xcolor}
\usepackage{capt-of}
\usepackage{makecell}
\usepackage{xspace}
\newcommand{\name}{MatRAG}

\begin{document}
\begin{frontmatter}
\title{A Matryoshka Hierarchical RAG for Efficient Multi-Hop Question Answering}

\author[1]{Gianluca Bonifazi}
\ead{g.bonifazi@univpm.it}

\author[1]{Christopher Buratti}
\ead{c.buratti@pm.univpm.it}

\author[1]{Michele Marchetti}
\ead{michele.marchetti@univpm.it}

\author[1]{Federica Parlapiano}
\ead{f.parlapiano@pm.univpm.it}

\author[1]{Giulia Quaglieri}
\ead{g.quaglieri@pm.univpm.it}

\author[2]{Davide Traini}
\ead{davide.traini@unimore.it}

\author[1]{Domenico Ursino}
\ead{d.ursino@univpm.it}

\author[1]{Luca Virgili\corref{cor1}}
\ead{luca.virgili@univpm.it}

\cortext[cor1]{Corresponding author}

\address[1]{Polytechnic University of Marche, Ancona, Italy}
\address[2]{University of Modena and Reggio Emilia, Modena, Italy}

\begin{abstract}
Retrieval-Augmented Generation (RAG) systems for multi-hop Question Answering (QA) must balance retrieval quality with computational cost. This cost is incurred during indexing time, through the use of expensive Knowledge Graphs (KGs) or Large Language Models (LLMs) to generate summaries, or during querying, through iterative LLM-driven retrieval. To reduce it while maintaining retrieval quality, we present \name{}, a hierarchical framework that combines RAG systems with Matryoshka Representation Learning (MRL). \name{} addresses both kinds of cost by aligning the semantic hierarchy of a clustering structure with the nested structure of MRL. Specifically, it organizes the corpus of documents into a Directed Acyclic Graph (DAG) of clusters with progressively coarser granularity. Each level is indexed by a lower Matryoshka dimension. \name{} pairs an iterative, top-down traversal of the DAG with an entity-driven mechanism that controls the hop budget and re-ranks candidates. We evaluated \name{} on three standard multi-hop QA benchmarks against seven representative baselines. \name{} outperforms its strongest competitors in terms of retrieval quality; furthermore, it reduces indexing costs by avoiding KG construction and LLM-based summarization, and lowers query-time costs through dimension-aware similarity.
\end{abstract}

\end{frontmatter}

\section{Introduction}
\label{sec:Introduction}

Retrieval-Augmented Generation (RAG) systems have become the dominant paradigm for grounding LLMs responses in external knowledge bases and reducing hallucinations~\cite{PeDrBo24, ViSiMo26}.

A significant challenge in RAG is multi-hop Question Answering (QA), where the correct answer requires aggregating information dispersed across several documents linked by entities or relationships~\cite{TaYa24}. A dense similarity between a query and a single document cannot capture chains of reasoning that span multiple documents, and flat indices do not exploit the hierarchical semantic structure of the corpus~\cite{Fan*24}. Recent literature has proposed two families of solutions to this problem. Graph-based approaches~\cite{Peng*25} construct an explicit Knowledge Graph (KG) of entities and relationships offline and explore it at query time via traversal~\cite{Wang*24-5}, Personalized PageRank~\cite{Gutierrez*25}, or LLM-based agents~\cite{Sun*24-2, Jin*24}. Hierarchical approaches organize the corpus into a multi-level index, enabling retrieval at multiple granularities~\cite{Tao*25, Sarthi*24}. However, both families incur significant computational costs that limit their scalability. Specifically, graph-based approaches incur offline costs through entity linking and relationship extraction, while hierarchical approaches incur offline costs through recursive LLM calls to summarize internal nodes~\cite{Sarthi*24}. Additionally, both families incur online costs through iterative LLM-driven planning \cite{Jiang*25-2} and expensive KG traversals \cite{Sun*24-2}.

We argue that these limitations could be overcome by using the nested representations provided by Matryoshka Representation Learning (MRL)~\cite{Kusupati*22}. MRL produces embeddings whose low-dimensional prefixes retain useful information. This allows us to obtain representations of different dimensionalities from a single embedding. We hypothesize that this property can be exploited in a hierarchical clustering structure, where progressively coarser levels require less fine-grained representations. Thus, the upper levels of the hierarchy can be indexed using shorter Matryoshka prefixes while  retaining the full embedding dimension for individual documents.

Starting from this intuition, in this paper we present \name{}, a Ma\-tryosh\-ka-indexed, hierarchical RAG framework designed for multi-hop QA. \name{} encodes each document offline by using a Matryoshka embedding and builds a hierarchical index through density-based clustering. This index is structured as a Directed Acyclic Graph (DAG). The leaves of the DAG store full-dimensional document embeddings, while the internal nodes store centroids at progressively lower Matryoshka dimensions towards the coarser levels. When a query is received, retrieval proceeds via a top-down traversal of the DAG. During this process, the query embedding is truncated to the dimension of the level being explored. This reduces the cost of similarity computations at the coarser levels. An entity-aware iterative mechanism controls how many documents are collected at each hop and how candidate documents are re-ranked. This replaces the iterative LLM calls of plan-then-retrieve approaches with a lightweight signal based on the entities found in the query and in the already retrieved context. 

We evaluated \name{} against seven representative RAG baselines on three multi-hop QA benchmarks, namely HotpotQA, 2WikiMultiHopQA, and MuSiQue. Our experimental campaign assessed retrieval effectiveness, answer quality, and computational efficiency. \name{} achieved the highest Exact Match (EM) and F1 scores across all three benchmarks, while consistently improving Recall@5 over the document-retrieval baselines. Moreover, it reduced the average response time compared to flat FAISS-based retrieval and substantially lowered the indexing costs compared to graph-based and hierarchical approaches.

The main contributions of this paper are:

\begin{itemize}

    \item A resolution-aligned indexing strategy that aligns MRL dimensions with index depth. This reduces the cost of query-time similarity at coarser levels while preserving retrieval accuracy.
    
    \item An iterative retrieval mechanism that does not rely on graph-based multi-hop traversal, but rather progressively expands the retrieved context through hierarchical search.
    
    \item An entity-aware re-ranking and drift-control mechanism, which uses entities to prioritize candidate nodes and prevents retrieval drift across iterations.

\end{itemize}

The rest of this paper is organized as follows: Section~\ref{sec:Related-Wok} reviews related literature. Section~\ref{sec:Proposed-Approach} introduces \name{}. Section~\ref{sec:experiments} reports our experimental results. Section~\ref{sec:Discussion} discusses findings and limitations. Finally, Section~\ref{sec:Conclusion} presents our conclusions and outlines directions for future work.

\section{Related Work}
\label{sec:Related-Wok}

This section describes related work and is divided into four subsections. In particular, Subsection~\ref{sub:RAG-Generation-Multi-Hop-QA} discusses Retrieval-Augmented Generation and multi-hop QA. Subsection~\ref{sub:Graph-Based-RAG} presents graph-based RAG approaches. Subsection~\ref{sub:Hierarchical-RAG} covers hierarchical RAG approaches. Finally, Subsection~\ref{sub:Matryoshka-Representation-Learning} introduces MRL for retrieval.

\subsection{Retrieval-Augmented Generation and Multi-Hop QA}
\label{sub:RAG-Generation-Multi-Hop-QA}

A RAG~\cite{Fan*24} incorporates external knowledge bases into LLM generation via an explicit retrieval step. The application of RAGs to multi-hop QA~\cite{TaYa24} ranges from single-pass retrieval to iterative pipelines that interleave retrieval and reasoning. Recent approaches have also explored LLM-based re-ranking strategies to improve the quality of retrieved evidence \cite{Song*26}. One such strategy involves decomposing the query into sub-steps and planning the subsequent retrieval~\cite{Jiang*25-2}. Instead, the strategy in~\cite{Zhang*25-3} uses explicit logic trees to generate more interpretable answers. While these strategies improve quality, they require repeated LLM calls at query time, and the cost increases with each additional hop. Other approaches improve retrieval by enriching the context used to identify relevant evidence \cite{Chu*26}. A known failure mode of the pipelines that append retrieved evidence to the query is query drift, which was originally studied in Pseudo-Relevance Feedback (PRF)~\cite{ZiKu08,Rocchio71}. In this mode, the expansion of the query with the top-ranked documents can shift the search away from the original information need. Recently, query drift has reemerged in dense retrieval~\cite{Li*23-4, Rathee*26} and iterative, multi-hop and graph-based RAG~\cite{Huang*26,Lau*26}. The classical mitigation strategy~\cite{Rocchio71}, which combines the original and expanded queries in the similarity computation, is the basis of our anchoring mechanism. 

\name{} fits into this framework while avoiding iterative LLM calls. Instead of relying on LLM-based retrieval planning, it uses entity counts to determine the retrieval budget at each iteration and entity overlap to re-rank candidate documents. Additionally, it mitigates query drift through an anchoring schedule that strengthens the original query as retrieval proceeds.

\subsection{Graph-Based RAG}
\label{sub:Graph-Based-RAG}

Graph-based approaches~\cite{Peng*25} construct a KG offline starting from the entities and relationships extracted from documents. These approaches use the KG's structure to guide retrieval when a query is submitted. The main distinction among these approaches is the granularity level with which the graph is built and queried. GraphRAG~\cite{Edge*24} operates at the community level, with communities detected by the Leiden algorithm~\cite{TrWaVa19}, and employs multi-level summaries for query-focused summarization. LightRAG~\cite{Guo*25-2} combines this global view with a local, entity-based view through dual-level retrieval. HippoRAG2~\cite{Gutierrez*25} operates at the entity level. It reframes retrieval as an associative memory problem and uses Personalized PageRank to identify relevant passages in one step. KGP~\cite{Wang*24-5} moves further away from entities by building a graph of passages with edges based on semantic and structural relationships traversed by an LLM-based agent. Another way of proceeding exploits the LLM as a reasoning engine to navigate the graph step by step, as in Think-on-Graph (ToG)~\cite{Sun*24-2} and Graph-CoT~\cite{Jin*24}, or to extract task-relevant substructures, as in SG-RAG~\cite{SaTuSa24} and KG\textsuperscript{2}RAG~\cite{Zhu*25-2}. These approaches perform well on multi-hop tasks but incur a significant indexing cost and depend on entity linking and relationship extraction, which are noisy processes by nature. 

\name{} only uses entities as a lightweight signal for budget control and re-ranking, and it does not construct an explicit KG.

\subsection{Hierarchical RAG}
\label{sub:Hierarchical-RAG}

Hierarchical RAG approaches organize the corpus into tree-like structures, in which retrieval operates at various granularity levels~\cite{Wang*26, PeLu27}. RAPTOR~\cite{Sarthi*24} builds the tree recursively, representing each internal node with an LLM-generated summary of its child nodes. Several approaches build on this paradigm by enriching the index with additional signals. For instance, SiReRAG~\cite{Zhang*25-4} indexes similar and related information together to support multi-hop reasoning, and~\cite{Huang*25} integrates hierarchical knowledge to bridge the gap between local and global contexts. Other works focus on how the hierarchy itself is constructed. For instance, ArchRAG~\cite{Wang*26} groups documents into communities with attributes, and TreeRAG~\cite{Tao*25} exploits the internal structure of lengthy documents to generate a hierarchy. Despite their differences, these approaches share two characteristics. First, they use a single embedding dimension across all tree levels. Second, they rely on LLM calls to derive representations of internal nodes. 

\name{} differs from these approaches in both respects. In fact, internal nodes are computed as centroids of their child nodes without LLM intervention. Additionally, MatRAG uses lower embedding dimensions for coarser clusters and progressively larger dimensions for finer-grained clusters, retaining full-dimensional embeddings only for individual documents.

\subsection{Matryoshka Representation Learning for Retrieval}
\label{sub:Matryoshka-Representation-Learning}

MRL~\cite{Kusupati*22} introduces a multi-scale loss that renders the prefixes of an embedding independently informative. Thus, a single model produces nested representations that are truncated at deployment time to balance cost and accuracy. This property has been combined with knowledge distillation for dense retrieval~\cite{Zhang*24-2}, applied to RAG in low-resource languages~\cite{NaSiCo24}, integrated into a hybrid retriever for general RAG~\cite{LiWa25}, used for similarity analysis on trajectory embeddings~\cite{PeGuGa25}, and leveraged for  interpretable, hierarchical clustering of multilingual news articles~\cite{HaDu25}. All these approaches use MRL to select one deployment dimension that balances cost and accuracy. However, they apply the chosen representation uniformly across the downstream task. 

Unlike them, \name{} proposes a structured use of MRL as an indexing strategy for hierarchical clustering, where progressively shorter prefixes are used at coarser levels of the hierarchy.
 
\section{Proposed Approach}
\label{sec:Proposed-Approach}

\begin{figure}[ht!]
    \centering
    \includegraphics[width=0.8\textwidth]{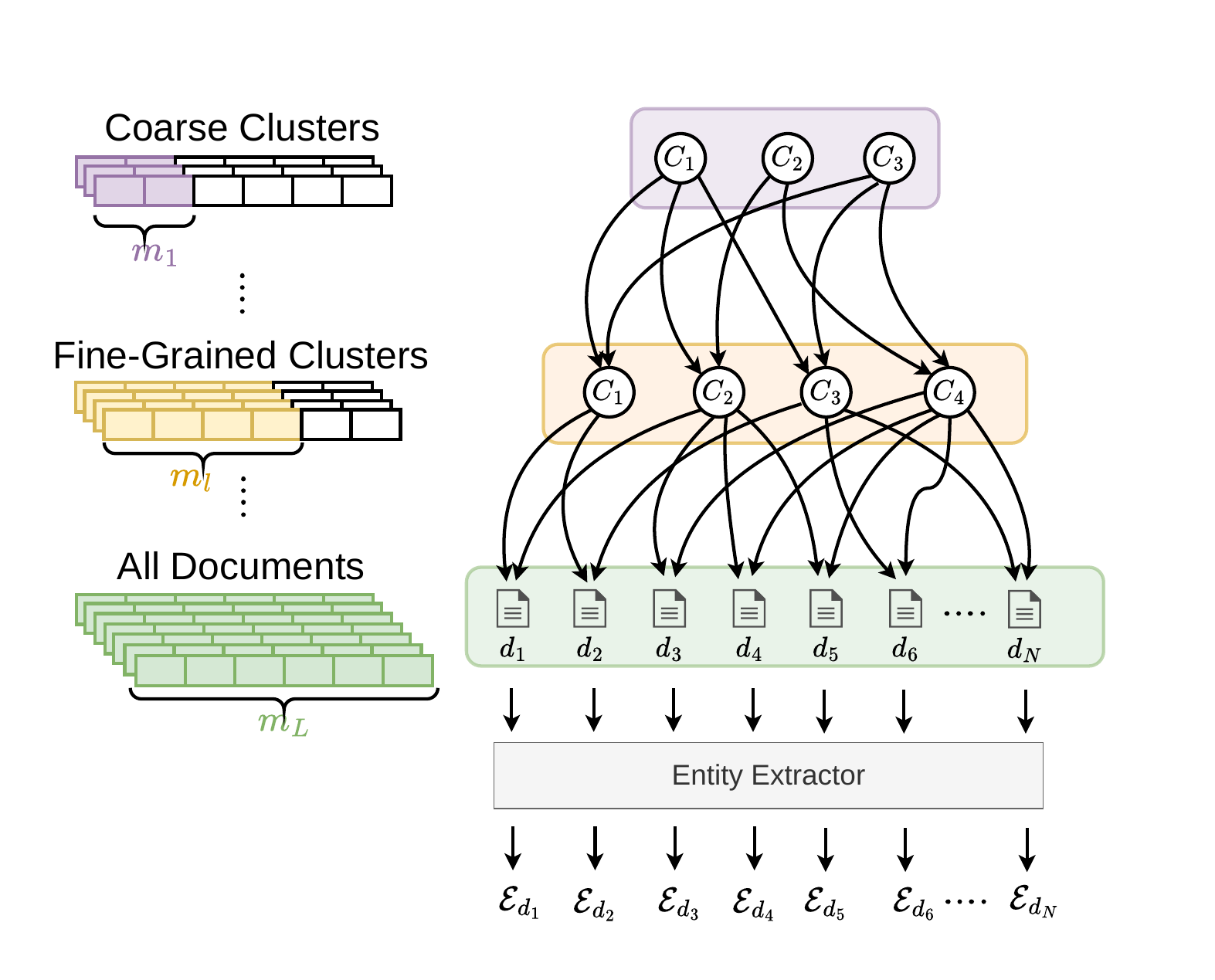}
    \caption{Hierarchical indexing (offline). The corpus is encoded using a Matryoshka embedding model and organized from the bottom up into a hierarchical DAG via HDBSCAN with overlapping cluster assignments. The internal nodes are progressively coarser cluster centroids stored at shorter Matryoshka prefixes. The leaf layer retains full-dimensional document embeddings. Named entities are extracted from each document and stored with the embeddings. This allows for retrieval using both semantic and entity-level information.}
    \label{fig:Overview-indexing}
\end{figure}

\begin{figure}[ht!]
    \centering
    \includegraphics[width=\textwidth]{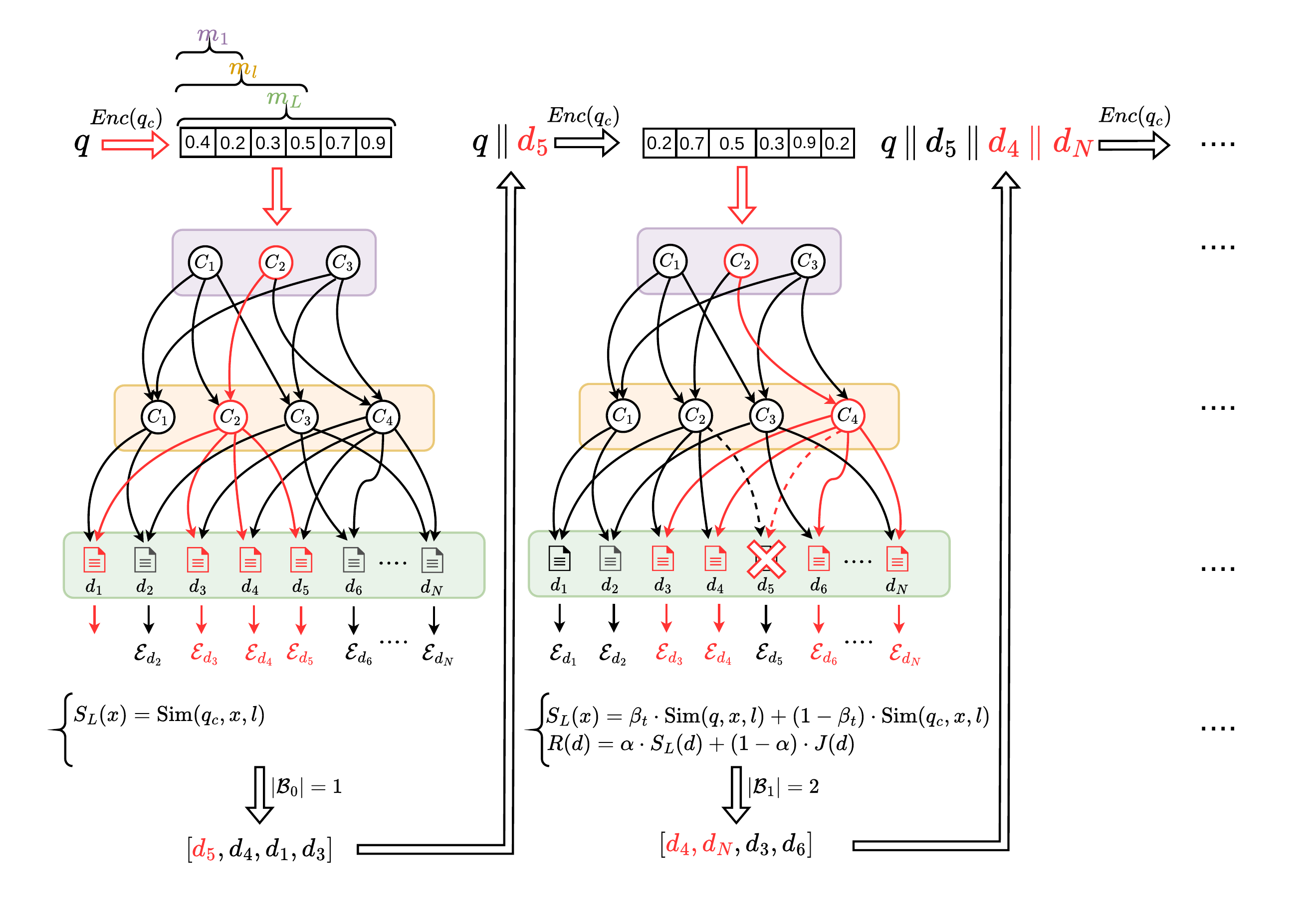}
    \caption{Entity-budgeted retrieval (online). Retrieval involves an iterative, top-down traversal of the hierarchy. Starting with the coarsest level, the query is scored against cluster centroids. At each level, the search space narrows to the most promising cluster until reaching the document level. The traversal score considers both similarity to the original query and similarity to the cumulative query. The latter incorporates previously retrieved documents to guide multi-hop reasoning while limiting query drift. At the leaves, candidates are re-ranked using a score that blends semantic similarity and entity overlap to prioritize documents that are both relevant and entity-consistent.}
    \label{fig:Overview-retrieval}
\end{figure}

In this section, we present \name{}, our hierarchical RAG framework designed for multi-hop QA. Its behavior consists of two phases. During the offline phase, \name{} encodes the corpus using a Matryoshka embedding model and builds a hierarchical index, in which internal nodes are stored at  progressively lower prefix dimensions. During the online phase, given a query, \name{} iteratively retrieves documents through a top-down traversal of the index and scores candidates with two complementary signals applied at different stages of the traversal. Figures \ref{fig:Overview-indexing} and \ref{fig:Overview-retrieval} illustrate the \name{} workflow.

\subsection{Hierarchical Indexing}
\label{sub:Hierarchical-Indexing}

Let $\mathcal{D} = \{d_1, \ldots, d_N\}$ be a corpus of documents. We encode each document $d_i \in \mathcal{D}$ using an MRL embedding model~\cite{Kusupati*22}, which produces a dense vector $e_i \in \mathbb{R}^{M}$. The MRL model organizes the information in $e_i$ along $L$ nested levels of granularity, such that, for any $l \in \{1, \ldots, L\}$, the first $m_l$ coordinates of $e_i$ form a self-contained representation of $d_i$ at the dimension $m_l$, with $m_1 < m_2 < \cdots < m_L = M$. Thus, the full vector $e_i$ contains $L$ nested representations of $d_i$, obtained by considering prefixes of increasing dimensionality. We compute each embedding $e_i$ once and store it at the full dimension $M$, since any lower-dimensional representation can be obtained as a prefix of it on the fly. We also extract a set of named entities $\mathcal{E}_i$ from each document $d_i \in \cal{D}$ using GLiNER~\cite{Zaratiana*24}, an open-schema NER model. We compute the entity sets once and store them so that we can efficiently access $\mathcal{E}_i$ for any document $d_i$ at retrieval time. Entity counts and overlaps allow us to control the iterative retrieval loop and re-rank candidate documents (see Section~\ref{sub:Entity-Budgeted-Retrieval}).

With the document embeddings and entity sets in place, we organize the corpus into a hierarchical index. This index exposes the semantic structure of $\mathcal{D}$ at multiple levels of granularity through hierarchical clustering. Specifically, we build a DAG whose nodes are partitioned into $L$ levels. 

We index the document level of the DAG by $l = L$, and the coarsest level by $l = 1$. The nodes at the level $L$ represent the documents in $\mathcal{D}$, while each level $l < L$ groups the nodes of the level $l + 1$ into clusters by applying HDBSCAN~\cite{Campello*15}. 

To capture the fact that a node can be semantically related to multiple clusters, we associate each node at the level $l + 1$ with its $p$ nearest clusters at the level $l$, as measured by the cosine similarity between the node and the cluster centroids. Therefore, each node has at most $p$ incoming edges, and the graph is acyclic because edges always link subsequent layers. Figure~\ref{fig:Overview-indexing} shows the structure of the DAG. The $p$-nearest assignment provides redundant coverage during top-down traversal. In fact, a node that semantically belongs to multiple clusters can be accessed from any of them, which makes the descent robust to suboptimal centroid choices at a given level.

We couple the clustering procedure with the storage of the internal nodes using the Matryoshka organization. Nodes at the level $L$ store document embeddings at the full dimension $M$, while every internal node at the level $l < L$ represents a centroid truncated at the reduced dimension $m_l$. The centroid of a cluster $\mathcal{C}$ at the level $l$ is computed as follows:

\begin{equation}
\mu_{\mathcal{C}}^{(l)} = \left( \frac{1}{|\mathcal{C}|} \sum_{c \in \mathcal{C}}
c \right)\Bigg|_{m_l}
\end{equation}

The elements of $\mathcal{C}$, which are already stored at the dimension $m_{l+1}$, are averaged directly. The symbol $|_{m_l}$ indicates that the resulting centroid is truncated to the dimension $m_l$ for storage.

This alignment is based on the idea that the coarser levels of the DAG aggregate semantically heterogeneous content into broad clusters. The discriminative information needed to rank candidates within these clusters is coarse and can be captured by shorter Matryoshka prefixes. In contrast, we reserve the full embedding dimension for the level $L$, where the final ranking of individual documents requires the most precise representation. 

\subsection{Entity-Budgeted Retrieval}
\label{sub:Entity-Budgeted-Retrieval}

We organize the retrieval procedure in the DAG as an iterative loop; each iteration of this loop collects a number of documents equal to the entities observed in the previous iteration. The idea is that the number of distinct entities in the query and the retrieved context serves as an indicator of the residual complexity of the multi-hop reasoning chain and defines the retrieval budget. The latter indicates the maximum number of documents to retrieve. Figure~\ref{fig:Overview-retrieval} illustrates this process. 

Let $\mathcal{G}$ be the hierarchical DAG built offline (Section~\ref{sub:Hierarchical-Indexing}), let $K$ be the total retrieval budget defined by the user, and let $q$ be the input query. The retrieval process is iterative and progressively expands the query context through entities extracted from previously retrieved documents. Algorithm~\ref{alg:loop} formalizes it.

During the first retrieval step ($t=1$), the process relies solely on the entities extracted from the input query. The function $\textsc{Retrieve}$ (Section~\ref{sub:Multi-Signal-Document-Scoring}) performs a top-down traversal of $\mathcal{G}$ and returns an initial set of retrieved documents, denoted as $\mathcal{D}_1$. These documents form the initial retrieved set $\mathcal{R}_1 = \mathcal{D}_1$. The entities extracted from the retrieved documents are accumulated into the entity set $\tilde{\mathcal{E}}_1$, along with the query entities. The subset of entities not present in the query defines the frontier entity set $\mathcal{B}_1$, which guides the next retrieval step.

For each subsequent iteration $t>1$, the retrieval process is guided by the entities discovered in the previous iteration. Let $q_c$ denote the cumulative query, obtained by concatenating the original query $q$ with the text of all documents retrieved thus far. At iteration $t$, $\textsc{Retrieve}$ traverses $\mathcal{G}$ using $q_c$ and re-ranks the candidate documents based on the entities in $\mathcal{B}_{t-1}$ according to the multi-signal document score (Section~\ref{sub:Multi-Signal-Document-Scoring}). Then, the process selects the top $k_t$ documents not contained in the cumulative retrieved set $\mathcal{R}_{t-1}$ and returns them as the new document set $\mathcal{D}_t$. $k_t = \min\left(K - \lvert R_{t-1}\rvert,\; \max\left(1,\lvert {\cal B}_{t-1}\rvert\right)\right)$ assures that the number of selected documents is no greater than $K$. The retrieved set is updated as $\mathcal{R}_t = \mathcal{R}_{t-1} \cup \mathcal{D}_t$. The entities extracted from $\mathcal{D}_t$ are accumulated into the cumulative entity set $\tilde{\mathcal{E}}_t$; the entities present in $\tilde{\mathcal{E}}_t$ and not in $\tilde{\mathcal{E}}_{t-1}$ define the new frontier set $\mathcal{B}_t$, which drives the next iteration. 

Finally, the retrieved set $\mathcal{R}_t$ is passed to an LLM together with $q$ to obtain the final answer.

\begin{algorithm}
\caption{Entity-Budgeted Retrieval Loop}
\label{alg:loop}
{\scriptsize
\begin{algorithmic}[1]
\Require $K$: a positive integer denoting the maximum number of documents to retrieve; $q$: a query; $\hat{\mathcal{E}}_q$: the entities of $q$; $\mathcal{G}$: a DAG
\State $\mathcal{R}_0 \gets \emptyset$, $\tilde{\mathcal{E}}_0 \gets
    \hat{\mathcal{E}}_q$, $\mathcal{B}_0 \gets \hat{\mathcal{E}}_q$, $t \gets 1$
\While{$|\mathcal{R}_{t-1}| < K$ \textbf{and} $\mathcal{B}_{t-1} \neq
    \emptyset$}
    \State $q_c \gets q \,\|\,
        \bigoplus_{d \in \mathcal{R}_{t-1}} d$
        \Comment{concatenate $q$ with the documents retrieved up to the iteration $t-1$}
    \State $k_t \gets \min\left(K - \lvert R_{t-1}\rvert,\; \max\left(1,\lvert \mathcal{B}_{t-1}\rvert\right)\right)$
    \Comment{number of documents to retrieve according to the remaining budget}
    \State $\mathcal{D}_t \gets \textsc{Retrieve}(\mathcal{G},
        q, q_c, \mathcal{R}_{t-1}, k_t)$
    \State $\tilde{\mathcal{E}}_t \gets \tilde{\mathcal{E}}_{t-1} \cup
        \big(\bigcup_{d \in \mathcal{D}_t} \mathcal{E}_d\big)$
        \Comment{accumulate entities from the retrieved documents}
    \State $\mathcal{R}_t \gets \mathcal{R}_{t-1} \cup \mathcal{D}_t$
    \State $\mathcal{B}_t \gets \tilde{\mathcal{E}}_t \setminus \tilde{\mathcal{E}}_{t-1}$
    \State $t \gets t + 1$
\EndWhile
\State $t \gets t - 1$
\State \textbf{return} $\mathcal{R}_t$
\end{algorithmic}
}
\end{algorithm}

\subsection{Multi-Signal Document Scoring}
\label{sub:Multi-Signal-Document-Scoring}

Multi-signal document scoring is performed by the $\textsc{Retrieve}$ function. As discussed in Section \ref{sub:Entity-Budgeted-Retrieval}, this function traverses ${\cal G}$ using $q_c$ and re-ranks the candidate documents according to the entities of $\mathcal{B}_{t-1}$. To perform this task, it ranks candidates using two complementary scores, i.e., $S_l(x)$ and $R(d)$. The traversal score $S_l(x)$ combines similarity to the original query (which acts as an anchor against query drift) with a contribution from the cumulative query (which provides context from documents retrieved thus far). Once the traversal reaches the level $L$, the re-ranking score $R(d)$ refines the ordering of the candidates. For this purpose, $R(d)$ incorporates an entity-overlap signal that rewards documents whose entities are coherent with those introduced at the previous iterations. Algorithm~\ref{alg:retrieve} details the $\textsc{Retrieve}$ function.

\begin{algorithm}[ht!]
\caption{\textsc{Retrieve} function}
\label{alg:retrieve}
{\scriptsize
\begin{algorithmic}[1]
\Require $\mathcal{G}$: a DAG; $q$: a query; $q_c$: a cumulative query; ${\cal R}_{t-1}$: a set of documents; $k_t$ a non-negative integer denoting the number of documents to retrieve at iteration $t$
\State $\mathcal{N} \gets \textsc{FirstLayer}(\mathcal{G})$
    \Comment{centroid vectors at level $1$}
\For{$l = 1$ \textbf{to} $L - 1$}
    \For{each centroid vector $e_i \in \mathcal{N}$}
        \State Compute $Sim(q, e_i, l)$ and $Sim(q_c, e_i, l)$ by applying Equation~\ref{eq:Sim}
        \State Compute $S_l(e_i)$ by applying Equation~\ref{eq:traversal_score}
    \EndFor
    \State $e_x \gets \arg\max_{e_i \in \mathcal{N}}\; S_l(e_i)$
    \State $\mathcal{N} \gets \textsc{Children}(e_x,\, \mathcal{G})$
        \Comment{vectors at level $l{+}1$ reachable from $e_x$}
\EndFor \Comment{At this point, $\mathcal{N}$ contains the embeddings of the candidate documents at level $l$}
\State $\mathcal{S} \gets \emptyset$
    \Comment{set of (documents, score) pairs}
\For{each $e_i \in \mathcal{N}$ such that $d_i \notin \mathcal{R}_{t-1}$}
    \State Compute $Sim(q, e_i, L)$ and $Sim(q_c, e_i, L)$ by applying Equation~\ref{eq:Sim}
    \State Compute $S_L(e_i)$ by applying Equation~\ref{eq:traversal_score}
    \State Compute $J(d_i)$ by applying Equation~\ref{eq:jaccard}
    \State Compute $R(d_i)$ by applying Equation~\ref{eq:rerank_score}
    \State $\mathcal{S} \gets \mathcal{S} \cup \{(d_i,\, R(d_i))\}$
\EndFor
\State \textbf{return} $\textsc{Top}\!\left(\mathcal{S},\;k_t\right)$
    \Comment{return the first $k_t$ documents in $\mathcal{S}$ ranked by $R(d_i)$}
\end{algorithmic}
}
\end{algorithm}

\subsubsection{Computation of the traversal score}
\label{subsub:Computation-Traversal-Score}

For each level $l$ of the DAG, the candidate nodes are ranked by a function $S_l(\cdot)$, which combines contributions from the original query $q$ and  the cumulative query $q_c$. In order to define $S_l(\cdot)$, we first introduce the level-aware similarity function as follows:

\begin{equation}
Sim(s, e_i, l) = \cos\!\left(e_s|_{m_l}, e_i\right)
\label{eq:Sim}
\end{equation}

Here, $s$ is a text, $e_s$ is its embedding, and $e_s|_{m_l}$ is the $m_l$-length prefix of $e_s$. The second argument of the function is the embedding $e_i$ stored at the level $l$ of the DAG. If $l$ is an internal level, then $e_i$ is the embedding of a centroid at $l$; instead, if $l$ is a leaf, then $e_i$ is the embedding of a document in ${\cal D}$. In both cases, $e_i$ is already stored as an $m_l$-length vector, so it does not require truncation. The similarity function $S_l(\cdot)$ returns a score that combines two contributions. The first contribution is the similarity of $e_i$ to the original query $q$, which anchors the search to the original information need ~\cite{Rocchio71}. The second contribution is the similarity of $e_i$ to the cumulative query $q_c$, which expands $q$ with the content of the documents retrieved thus far via  PRF~\cite{ZiKu08}:

\begin{equation}
S_l(e_i) = \beta_t \cdot Sim(q, e_i, l) + (1 - \beta_t) \cdot \mathrm{Sim}(q_c, e_i, l)
\label{eq:traversal_score}
\end{equation}

where the anchoring coefficient $\beta_t$ is defined as follows:

\begin{equation}
\beta_t = \frac{|\mathcal{R}_{t-1}|}{K}
\label{eq:beta_t}
\end{equation}

In the first iteration, $q_c = q$ and $\beta_t = 0$, which reduces $S_l(e_i)$ to $Sim(q, e_i, l)$. As documents are collected, $q_c$ incorporates their content progressively and may drift away from the original information need. Concurrently, $\beta_t$ increases, shifting the weight toward $q$ and preventing semantic drift.

The traversal begins at the coarsest level (i.e., $l = 1$) and continues downward. At each level $l$, we apply the function $S_l(\cdot)$ to each centroid at $l$. Then, we select the centroid $e_x$ of $l$ with the highest value of $S_l(\cdot)$. Afterwards, we restrict the candidate set to the nodes at the level $l + 1$ that are connected to the node corresponding to $e_x$. Due to the $p$-nearest assignment introduced in Section~\ref{sub:Hierarchical-Indexing}, each node at the level $l + 1$ can be reached from up to $p$ nodes at the level $l$. Thus, the traversal can tolerate suboptimal centroid choices without losing relevant candidates. 

We repeat this process until the level $L$ is reached. At this level, the candidates are documents in $\mathcal{D}$, and their embeddings are stored at the full dimension $M$. Since each level $l < L$ operates at a lower dimension $m_l<M$ rather than the full dimension $M$, and since the candidate set is progressively restricted as the traversal proceeds, the similarity computations at the coarser levels are cheaper and performed over a smaller set of candidates. This reduces the overall cost compared to a flat search over the full corpus at the dimension $M$.

\subsubsection{Computation of the re-ranking score}
\label{subsub:Computation-Reranking-Score}
After traversing to the level $L$, we re-rank the candidates using the set of new entities $\mathcal{B}_{t-1}$ (see Section~\ref{sub:Entity-Budgeted-Retrieval}). To this end, we use the Jaccard coefficient to measure the overlap between the set of entities $\mathcal{E}_i$ of a candidate document $d_i$ and the set of new entities $\mathcal{B}_{t-1}$ as follows:

\begin{equation}
J(d_i) = \frac{|\mathcal{B}_{t-1} \cap \mathcal{E}_i|} {|\mathcal{B}_{t-1} \cup \mathcal{E}_i|}
\label{eq:jaccard}
\end{equation}

The final score $R(d_i)$ of a document $d_i$ at the level $L$ combines semantic similarity and entity overlap as follows:

\begin{equation}
R(d_i) = \alpha \cdot S_L(e_i) + (1-\alpha) \cdot J(d_i)
\label{eq:rerank_score}
\end{equation}

where $\alpha$ controls the relative importance of the entity signal. The Jaccard coefficient ranges in the real interval $[0, 1]$ and is therefore stable with respect to the scale of cosine similarity. The $\textsc{Retrieve}$ function returns the top documents, ranked by $R(d_i)$, within the budget $K$ defined in Section~\ref{sub:Entity-Budgeted-Retrieval}. These documents form the set ${\cal D}_t$ and are then accumulated into $\mathcal{R}_t$ by the retrieval loop.

\section{Experimental Campaign}
\label{sec:experiments}

This section presents an experimental evaluation of \name{}. Specifically, Subsection~\ref{sub:Experimental-Setup} describes the experimental setup. Subsection~\ref{sub:Hyperparameter-Analysis} examines the impact of the main hyperparameters. Subsection~\ref{sub:Comparison-Results} reports the main results. Finally, Subsection~\ref{sub:Ablation-Analysis} presents an ablation study that evaluates the contribution of each component of \name{}.

\subsection{Experimental Setup}
\label{sub:Experimental-Setup}

In this section, we describe the setup used during our experimental campaign. Specifically, we focus on the datasets used, the baselines selected for comparison, some implementation choices for \name{}, and the evaluation metrics.

\subsubsection{Datasets}
\label{subsub:Datasets}

We evaluated \name{} on three standard multi-hop QA benchmarks, namely HotpotQA~\cite{Yang*18}, 2WikiMultiHopQA (2Wiki)~\cite{Ho*20}, and MuSiQue~\cite{Trivedi*22}. Each dataset requires aggregating evidence from multiple documents to answer a question, though datasets differed in the type of reasoning involved. In fact, HotpotQA covers bridge and comparison questions, 2Wiki focuses on compositional reasoning over Wikipedia, and MuSiQue is designed to resist shortcut-based retrieval. All three benchmarks provide gold answers, i.e., reference strings against which generated answers are evaluated. The benchmarks also provide gold supporting documents, which are subsets of the corpus documents containing the evidence necessary to answer each question. We used these documents to compute retrieval metrics. 

The document corpora contain $9,811$ documents for HotpotQA, $6,119$ documents for 2Wiki, and $11,656$ documents for MuSiQue. While previous studies typically relied on a single sample of $1,000$ questions \cite{Gutierrez*24, Press*23}, we drew five independent samples of $1,000$ questions each to reduce variability in question selection. We evaluated all methods on the same five samples, and we report the mean and standard deviation of each metric across them. We selected hyperparameter values using a separate validation set of 500 questions per benchmark that was disjoint from the test split.

\subsubsection{Baselines}
\label{subsub:Baselines}

We compared \name{} with seven representative RAG systems that span the main families of approaches discussed in
Section~\ref{sec:Related-Wok}. As a reference for flat retrieval, we included NaiveRAG, which performs dense retrieval over a FAISS\footnote{\url{https://faiss.ai/cpp_api/struct/structfaiss_1_1IndexFlatIP.html}} flat index.
As a representative of the hierarchical family, we included RAPTOR~\cite{Sarthi*24}. The remaining baselines covered graph-based and reasoning-augmented approaches; they are: GraphRAG~\cite{Edge*24}, LightRAG~\cite{Guo*25-2},
HippoRAG2~\cite{Gutierrez*25}, KGP~\cite{Wang*24-5}, and ToG~\cite{Sun*24-2}. For all baselines, we followed the original implementations and adopted the hyperparameter settings reported in the papers introducing them. To ensure a fair comparison, all baselines used the same generative model (\texttt{gemma3:27b-it-qat}\footnote{\url{https://ollama.com/library/gemma3:27b-it-qat}} via Ollama) for answer generation.

\subsubsection{Implementation Details}
\label{subsub:Implementation-Details}

All systems that require dense retrieval, including \name{}, use the nomic-embed-text-v1.5~\cite{Nussbaum*25} as the embedding model. This model natively supports MRL at the following dimensions: $\{64, 128, 256, 512, 768\}$. We adopted it as the Matryoshka dimension schedule for \name{}, setting $L = 5$ and $M = 768$. We stored the five levels of the DAG at the corresponding dimensions. Documents at the leaves maintained the full dimension $M = 768$, while internal clusters used progressively shorter prefixes toward the coarsest level, which was stored at $m_1 = 64$. We selected the values of the total retrieval budget $K$, the similarity weight $\alpha$, and the DAG overlapping factor $p$ via the hyperparameter analysis reported in Section~\ref{sub:Hyperparameter-Analysis}. In particular, based on this analysis, we set $K = 10$, $p = 2$, and $\alpha = 0.5$ consistently across all datasets. We conducted all experiments on a server equipped with an NVIDIA A100 GPU (40~GB VRAM) and 128~GB of system RAM. The interested reader can find the implementation code at the following link: \url{https://anonymous.4open.science/r/MatRAG}.

\subsubsection{Metrics}
\label{subsub:Metrics}

We evaluated the quality of the answers using Exact Match (EM) and token-level F1, computed against the gold answers~\cite{Yang*18,Ho*20,Trivedi*22}. To measure retrieval quality independently of the generative step, as in previous studies~\cite{Gutierrez*25,Gutierrez*24}, we employed Recall@2 (R@2) and Recall@5 (R@5). For each question, R@2 (resp., R@5) measures the number of gold supporting documents found among the top two (resp., five) retrieved documents divided by the total number of gold supporting documents for that question. We averaged these values across all questions in a dataset. We only computed these retrieval metrics for \name{}, HippoRAG2, LightRAG, and NaiveRAG, as these approaches retrieve documents directly from the corpus. The remaining baselines retrieve units not directly comparable to corpus documents. These include the node summaries of RAPTOR, the community summaries of GraphRAG, the passages of KGP (which have a different level of detail than the gold supporting documents), and the KG entities and relationships of ToG. For these systems, the gold document overlap required by R@2 and R@5 was undefined.

In addition to assessing retrieval and answer quality, we evaluated the efficiency of each system based on indexing time (Idx), response time (Res), and LLM context size (Tok). Indexing time is the amount of time, in seconds, required to build the index from the corpus. Response time is the average time, in seconds, for each retrieval and answer generation. LLM context size is the average number of tokens passed to the generative model per question.

\subsection{Hyperparameter Analysis}
\label{sub:Hyperparameter-Analysis}

In this section, we describe the analyses we performed to tune the hyperparameter values. To avoid tuning on the test data, we ran our analyses on a separate validation set of 500 questions for each dataset. We sampled this set disjointedly from the questions used for the test data (see Section \ref{sub:Experimental-Setup}). All reported values are means over this validation set.

First, we evaluated the effect of the retrieval budget $K$ by considering two values of this parameter: $K=5$ and $K=10$. The results are reported in Table~\ref{tab:hp-max-ctx-chunks}. In each column of this and the following tables, the optimal value is in bold, and the suboptimal value is underlined. An upward-pointing (resp., downward-pointing) arrow next to a metric indicates that its optimal value is the highest (resp., lowest). The analysis of this table shows that increasing $K$ from 5 to 10 substantially improves the EM and F1 scores across all datasets. As for EM, there are gains of 16.34\% on HotpotQA, 19.25\% on 2Wiki and 21.36\% on MuSiQue. Regarding F1, we observe  improvements of 8.56\% on HotpotQA, 19.34\% on 2Wiki and 21.45\% on MuSiQue. Conversely, we observe an increase in Tok, which increased by 108.00\% on HotpotQA, 104.03\% on 2Wiki, and 102.26\% on MuSiQue. These results confirm that multi-hop QA accuracy improves with larger contexts because additional documents help cover longer reasoning chains. However, this comes at the expense of efficiency. We chose to prioritize accuracy over efficiency and set $K$ to 10. At the same time, we did not set $K$ to a greater value so as to not dramatically worsen efficiency.

\begin{table}[!ht]
\scriptsize
\centering
\caption{Values of Exact Match (EM), token-level F1 (F1) and LLM context size (Tok) obtained by \name{} on the three benchmarks for $K=5$ and $K=10$. The values represent the results computed on the validation set consisting of 500 different questions. For each column, the optimal value is shown in bold.}
\label{tab:hp-max-ctx-chunks}
\begin{tabular}{l|ccc ccc ccc}
\toprule
 & \multicolumn{3}{c}{HotpotQA} & \multicolumn{3}{c}{2Wiki} & \multicolumn{3}{c}{MuSiQue} \\
\cmidrule(lr){2-4} \cmidrule(lr){5-7} \cmidrule(lr){8-10}
$K$ & EM $\uparrow$ & F1 $\uparrow$  & Tok $\downarrow$ & EM $\uparrow$ & F1 $\uparrow$ & Tok $\downarrow$ & EM $\uparrow$ & F1 $\uparrow$ & Tok $\downarrow$ \\
\midrule
5  & 51.40 & 65.17 & \textbf{325} & 42.60 & 49.85 & \textbf{298} & 20.60 & 29.97 & \textbf{354} \\[0.5ex]
10 & \textbf{59.80} & \textbf{70.75} & 676 & \textbf{50.80} & \textbf{59.49} & 608 & \textbf{25.00} & \textbf{36.40} & 716\\
\bottomrule
\end{tabular}
\end{table}

After analyzing $K$, we proceeded to analyze the impact of $p$, which controls the number of cluster assignments per node in the DAG. Table~\ref{tab:hp-top-k-clusters-per-doc} shows the obtained results. From the analysis of this table, we can see that the best performance is consistently achieved with $p = 2$, improving F1 (resp., EM) by 3.04\% (resp., 5.65\%) on HotpotQA, 4.85\% (resp., 7.63\%) on 2Wiki, and 8.20\% (resp., 9.65\%) on MuSiQue compared to $p = 1$. Increasing $p$ beyond 2 does not provide further gains and, in some cases, it slightly degrades performance (e.g., we can observe a 0.11\% decrease in F1 on HotpotQA when increasing $p$ from 2 to 3). Analogous conclusions can be drawn for R@2 and R@5. These results suggest that a limited amount of redundancy is sufficient to ensure robust traversal without introducing excessive noise.

\begin{table}[!ht]
\scriptsize
\centering
\caption{Values of EM, F1, Recall@2 (R@2) and Recall@5 (R@5) obtained by \name{} on the three benchmarks for $p$ ranging from 1 to 4. The values represent the results computed on the validation set consisting of 500 different questions. For each column, the optimal value is shown in bold.}
\label{tab:hp-top-k-clusters-per-doc}
\begin{tabular}{l|ccccccccc}
\toprule
 & \multicolumn{3}{c}{HotpotQA} & \multicolumn{3}{c}{2Wiki} & \multicolumn{3}{c}{MuSiQue} \\
\cmidrule(lr){2-4} \cmidrule(lr){5-7} \cmidrule(lr){8-10}
 $p$ & EM $\uparrow$ & F1 $\uparrow $& \begin{tabular}{@{}c@{}}R@2 $\uparrow$\\R@5 $\uparrow$ \end{tabular} & EM $\uparrow$ & F1 $\uparrow$& \begin{tabular}{@{}c@{}}R@2 $\uparrow $\\R@5 $\uparrow$ \end{tabular} & EM $\uparrow$ & F1 $\uparrow$ & \begin{tabular}{@{}c@{}}R@2 $\uparrow$ \\R@5 $\uparrow$ \end{tabular} \\
\midrule
1 & 56.60 & 68.66 & \begin{tabular}{@{}c@{}}73.80\\88.70\end{tabular} & 47.20 & 56.74 & \begin{tabular}{@{}c@{}}63.45\\78.35\end{tabular} & 22.80 & 33.64 & \begin{tabular}{@{}c@{}}42.60\\55.87\end{tabular} \\[0.5ex]
\cmidrule(lr){2-10}
2 & \textbf{59.80} & \textbf{70.75} & \begin{tabular}{@{}c@{}}\textbf{75.80}\\\textbf{91.10}\end{tabular} & \textbf{50.80} & \textbf{59.49} & \begin{tabular}{@{}c@{}}\textbf{66.05}\\\textbf{81.85}\end{tabular} & \textbf{25.00} & \textbf{36.40} & \begin{tabular}{@{}c@{}}\textbf{43.78}\\ \textbf{57.40}\end{tabular} \\[0.5ex]
\cmidrule(lr){2-10}
3 & 59.60 & 70.67 & \begin{tabular}{@{}c@{}}75.60\\90.30\end{tabular} & 50.00 & 59.19 & \begin{tabular}{@{}c@{}}65.35\\80.25\end{tabular} & 24.00 & 35.66 & \begin{tabular}{@{}c@{}}42.92\\57.22\end{tabular} \\[0.5ex]
\cmidrule(lr){2-10}
4 & 59.00 & 70.36 & \begin{tabular}{@{}c@{}}74.20\\89.50\end{tabular} & 49.00 & 58.19 & \begin{tabular}{@{}c@{}}65.05\\79.75\end{tabular} & 23.20 & 33.52 & \begin{tabular}{@{}c@{}}42.37\\56.98\end{tabular} \\
\bottomrule
\end{tabular}
\end{table}

Finally, we examined the effect of $\alpha$, which regulates the contribution of the entity-aware re-ranking. Table~\ref{tab:hp-alpha-used} shows the obtained results. It reveals that the best overall results are obtained at $\alpha = 0.50$, which provides a balanced combination of semantic similarity and entity overlap. Relying only on similarity-based ranking ($\alpha=1.00$) returns poorer performance. In fact, setting $\alpha=0.50$ improves F1 (resp., EM) by 4.63\% (resp., 3.10\%) on HotpotQA, 11.59\% (resp., 10.92\%) on 2Wiki, and 22.02\% (resp., 34.41\%) on MuSiQue, compared to setting $\alpha$ to 1.00. Relying exclusively on entity overlap ($\alpha = 0.00$) also degrades performance. In fact, setting $\alpha$ to 0.50 improves F1 (resp., EM) by 4.78\% (resp., 7.17\%) on HotpotQA, 2.13\% (resp., 0.79\%) on 2Wiki, and 3.79\% (resp., 4.16\%) on MuSiQue, compared to setting $\alpha$ to 0.00. When $\alpha$ is set to 0.25 (resp., 0.75) rather than 0.00 (resp., 1.00), results are sometimes better. However, they continue to be worse than when $\alpha$ is set to 0.50. R@2 and R@5 follow the same trends. Overall, these results suggest that neither semantic similarity nor entity overlap is sufficient on its own and that a balanced combination of the two, with $\alpha$ set to $0.50$, provides the most robust performance across the three benchmarks.

\begin{table}[!ht]
\scriptsize
\centering
\caption{Values of EM, F1, R@2 and R@5 obtained by \name{} on the three benchmarks for $\alpha$ set to 0.00, 0.25, 0.50, 0.75, and 1.00. The values represent the results computed on the validation set consisting of 500 different questions. For each column, the optimal value is shown in bold.}
\label{tab:hp-alpha-used}
\begin{tabular}{l|ccccccccc}
\toprule
 & \multicolumn{3}{c}{HotpotQA} & \multicolumn{3}{c}{2Wiki} & \multicolumn{3}{c}{MuSiQue} \\
\cmidrule(lr){2-4} \cmidrule(lr){5-7} \cmidrule(lr){8-10}
 $\alpha$ & EM $\uparrow $& F1 $\uparrow $& \begin{tabular}{@{}c@{}}R@2 $\uparrow$ \\R@5$\uparrow$ \end{tabular} & EM $\uparrow$ & F1 $\uparrow$ & \begin{tabular}{@{}c@{}}R@2 $\uparrow$ \\R@5 $\uparrow$ \end{tabular} & EM $\uparrow$ & F1 $\uparrow$ & \begin{tabular}{@{}c@{}}R@2 $\uparrow$ \\R@5 $\uparrow$ \end{tabular} \\
\midrule
0.00 & 55.80 & 67.52 & \begin{tabular}{@{}c@{}}74.70\\86.80\end{tabular} & 50.40 & 58.25 & \begin{tabular}{@{}c@{}}64.70\\81.15\end{tabular} & 24.00 & 35.07 & \begin{tabular}{@{}c@{}}\textbf{43.80}\\56.70\end{tabular} \\[0.5ex]
\cmidrule(lr){2-10}
0.25 & 57.20 & 68.48 & \begin{tabular}{@{}c@{}}74.30\\90.30\end{tabular} & 50.20 & 58.79 & \begin{tabular}{@{}c@{}}65.80\\80.70\end{tabular} & 23.40 & 34.79 & \begin{tabular}{@{}c@{}}43.03\\56.22\end{tabular} \\[0.5ex]
\cmidrule(lr){2-10}
0.50 & \textbf{59.80} & \textbf{70.75} & \begin{tabular}{@{}c@{}}75.80\\\textbf{91.10}\end{tabular} & \textbf{50.80} & \textbf{59.49} & \begin{tabular}{@{}c@{}}\textbf{66.05}\\\textbf{81.85}\end{tabular} & \textbf{25.00} & \textbf{36.40} & \begin{tabular}{@{}c@{}}43.78\\\textbf{57.40}\end{tabular}\\[0.5ex]
\cmidrule(lr){2-10}
0.75 & 58.80 & 69.81 & \begin{tabular}{@{}c@{}}\textbf{76.40}\\90.20\end{tabular} & 48.40 & 56.74 & \begin{tabular}{@{}c@{}}64.70\\76.60\end{tabular} & 19.80 & 30.20 & \begin{tabular}{@{}c@{}}43.70\\55.75\end{tabular} \\[0.5ex]
\cmidrule(lr){2-10}
1.00 & 58.00 & 67.62 & \begin{tabular}{@{}c@{}}74.30\\88.90\end{tabular} & 45.80 & 53.31 & \begin{tabular}{@{}c@{}}63.45\\73.80\end{tabular} & 18.60 & 29.83 & \begin{tabular}{@{}c@{}}42.61\\53.37\end{tabular} \\
\bottomrule
\end{tabular}
\end{table}

\subsection{Comparison Results}
\label{sub:Comparison-Results}

We first computed the retrieval performance of \name{} and the baselines. The results are reported in Table~\ref{tab:Comparison-Results-recall}. In each column of this and the following tables, the optimal value is in bold, and the suboptimal value is underlined. The analysis of Table~\ref{tab:Comparison-Results-recall} shows that \name{} consistently achieves optimal results across all benchmarks in most settings. On HotpotQA, it obtains the highest R@2 and R@5 scores, surpassing the strongest baselines. Specifically, it outperforms NaiveRAG by 2.98\% on R@2 and LightRAG by 5.99\% on R@5. On 2Wiki, \name{} achieves the best R@5 score, outperforming HippoRAG2 by 1.89\%. However, its R@2 is slightly below that of NaiveRAG, with a difference of 0.51\%. On the more challenging MuSiQue benchmark, \name{} delivers the strongest improvements, surpassing NaiveRAG, the best baseline, by 14.48\% on R@2 and 13.37\% on R@5. These results suggest that \name{} is particularly effective in complex, multi-hop retrieval scenarios.

\begin{table}[ht!]
\centering
\scriptsize
\caption{Retrieval performance on the three multi-hop benchmarks measured by Recall@2 (R@2) and Recall@5 (R@5). The results are reported only for systems that retrieve corpus documents directly. The values represent the means and standard deviations computed over 5 samples, each of which contains 1,000 different questions. For each column, the optimal value is shown in bold and the suboptimal value is underlined.}
\label{tab:Comparison-Results-recall}
\begin{tabular}{lcccccc}
\toprule
 & \multicolumn{2}{c}{HotpotQA} & \multicolumn{2}{c}{2Wiki} & \multicolumn{2}{c}{MuSiQue} \\
\cmidrule(r){2-3} \cmidrule(r){4-5} \cmidrule(r){6-7}
 & R@2 $\uparrow$ & R@5 $\uparrow$ & R@2 $\uparrow$ & R@5 $\uparrow$ & R@2 $\uparrow$ & R@5 $\uparrow$\\
\midrule
HippoRAG2 & 56.40\tiny{$\pm$.42} & 82.95\tiny{$\pm$.57} & 62.50\tiny{$\pm$.86} & \underline{77.80\tiny{$\pm$.31}} & 34.60\tiny{$\pm$.44} & 50.30\tiny{$\pm$.39} \\
LightRAG & 68.05\tiny{$\pm$.37} & \underline{84.25\tiny{$\pm$.29}} & 58.90\tiny{$\pm$.82} & 68.70\tiny{$\pm$.97} & 34.80\tiny{$\pm$.46} & 48.75\tiny{$\pm$.40} \\
NaiveRAG & \underline{70.55\tiny{$\pm$.54}} & 84.10\tiny{$\pm$.28} & \textbf{63.40\tiny{$\pm$.58}} & 69.55\tiny{$\pm$.85} & \underline{37.50\tiny{$\pm$.43}} & \underline{50.35\tiny{$\pm$.37}} \\
\midrule
\name{} & \textbf{72.65\tiny{$\pm$.89}} & \textbf{89.30\tiny{$\pm$.46}} & \underline{63.08\tiny{$\pm$.65}} & \textbf{79.27\tiny{$\pm$.81}} & \textbf{42.93\tiny{$\pm$.40}} & \textbf{57.08\tiny{$\pm$.33}} \\
\bottomrule
\end{tabular}
\end{table}

Next, we analyzed whether these improvements in retrieval translate into better performance in downstream QA. To this end, we computed the EM and F1 scores. The results are reported in Table~\ref{tab:Comparison-Results-emf1}. As can be seen from this table, \name{} achieves the best performance across all datasets and metrics. On HotpotQA, it improves upon NaiveRAG by 14.04\% in EM and LightRAG by 10.68\% in F1. On 2Wiki, the gains over the strongest baseline, HippoRAG2, are 5.71\% in EM and 14.89\% in F1. Finally, on MuSiQue, \name{} achieves the highest scores again, improving upon HippoRAG2 by 4.58\% in EM and 12.94\% in F1. Overall, these results demonstrate that improvements in retrieval quality lead to better QA performance, particularly in more challenging multi-hop settings.

\begin{table}[ht!]
\centering
\caption{QA performance on the three multi-hop benchmarks, measured by EM and F1 computed against the gold answers. The values represent the means and standard deviations computed over 5 samples, each of which contains 1,000 different questions. For each column, the optimal value is shown in bold and the suboptimal value is underlined.}
\label{tab:Comparison-Results-emf1}
\scriptsize
\begin{tabular}{lrrrrrr}
\toprule
 & \multicolumn{2}{c}{HotpotQA} & \multicolumn{2}{c}{2Wiki} & \multicolumn{2}{c}{MuSiQue} \\
\cmidrule(r){2-3} \cmidrule(r){4-5} \cmidrule(r){6-7}
 & EM $\uparrow$ & F1 $\uparrow$ & EM $\uparrow$ & F1 $\uparrow$ & EM $\uparrow$ & F1 $\uparrow$ \\
\midrule
HippoRAG2 & 52.40\tiny{$\pm$.58} & 64.10\tiny{$\pm$.62} & \underline{49.85\tiny{$\pm$.54}} & \underline{55.20\tiny{$\pm$.73}} & \underline{26.20\tiny{$\pm$.80} }& \underline{33.55\tiny{$\pm$.39}} \\
RAPTOR & 39.50\tiny{$\pm$.81} & 55.35\tiny{$\pm$.93} & 32.10\tiny{$\pm$.74} & 37.90\tiny{$\pm$.41} & 14.15\tiny{$\pm$.48} & 23.85\tiny{$\pm$.93} \\
KGP & 23.20\tiny{$\pm$.66} & 33.10\tiny{$\pm$.44} & 12.05\tiny{$\pm$.89} & 13.80\tiny{$\pm$.70} & 11.75\tiny{$\pm$.82} & 18.90\tiny{$\pm$.61} \\
ToG & 21.30\tiny{$\pm$.72} & 27.45\tiny{$\pm$.85} & 15.20\tiny{$\pm$.91} & 17.75\tiny{$\pm$.76} & 6.40\tiny{$\pm$.69} & 9.25\tiny{$\pm$.59} \\
GraphRAG & 30.65\tiny{$\pm$.63} & 40.60\tiny{$\pm$.78} & 8.25\tiny{$\pm$.56} & 9.10\tiny{$\pm$.51} & 1.10\tiny{$\pm$.32} & 2.30\tiny{$\pm$.38} \\
LightRAG & 52.15\tiny{$\pm$.81} & \underline{66.55\tiny{$\pm$.69}} & 45.35\tiny{$\pm$.42} & 50.80\tiny{$\pm$.37} & 21.85\tiny{$\pm$.93} & 31.80\tiny{$\pm$.41} \\
NaiveRAG & \underline{53.05\tiny{$\pm$.38}} & 64.50\tiny{$\pm$.95} & 44.10\tiny{$\pm$.32} & 48.15\tiny{$\pm$.49} & 20.65\tiny{$\pm$.39} & 29.55\tiny{$\pm$.30} \\
\midrule
\name{} & \textbf{60.50\tiny{$\pm$.37}} & \textbf{73.66\tiny{$\pm$.29}} & \textbf{52.70\tiny{$\pm$.35}} & \textbf{63.42\tiny{$\pm$.31}} & \textbf{27.40\tiny{$\pm$.38}} & \textbf{37.89\tiny{$\pm$.34}} \\
\bottomrule
\end{tabular}
\end{table}

Afterward, we evaluated the indexing time for each benchmark. The results are shown in Table~\ref{tab:indexing_time}. As reported in this table, \name{} is faster than most graph-based and structured approaches. Compared to HippoRAG2, it reduces indexing time by over 99.60\% across all datasets. It also outperforms other methods, such as GraphRAG and LightRAG, by several orders of magnitude. However, it remains slower than NaiveRAG, though it achieves better retrieval quality on most metrics. This efficiency gain primarily stems from the fact that many competing approaches (e.g., GraphRAG and RAPTOR) require expensive document summarization steps during graph construction, and that others (e.g., HippoRAG2 and LightRAG) necessitate a complex KG construction phase. In contrast, \name{} relies on simple cluster-level node representations obtained through node averaging, thus avoiding costly preprocessing.

\begin{table}[ht!]
\scriptsize
\centering
\caption{Indexing time (Idx) in seconds across the three benchmarks. The values represent the means and standard deviations computed over 5 independent runs of the indexing procedure on the full corpus of each benchmark. For each column, the optimal value is shown in bold, and the suboptimal value is underlined.}
\label{tab:indexing_time}
\begin{tabular}{lrrr}
\toprule
 & HotpotQA & 2Wiki & MuSiQue \\
\midrule
HippoRAG2 & 312,441\tiny{$\pm$1,842} & 123,187\tiny{$\pm$1,253} & 321,905\tiny{$\pm$1,976} \\
RAPTOR & 28,763\tiny{$\pm$412} & 16,724\tiny{$\pm$318} & 36,204\tiny{$\pm$487} \\
KGP & 1,941\tiny{$\pm$63} & 1,478\tiny{$\pm$51} & 4,213\tiny{$\pm$74} \\
ToG & 141,820\tiny{$\pm$934} & 69,340\tiny{$\pm$721} & 148,573\tiny{$\pm$1,102} \\
GraphRAG & 661,392\tiny{$\pm$2,341} & 75,614\tiny{$\pm$843} & 197,840\tiny{$\pm$1,587} \\
LightRAG & 728,471\tiny{$\pm$2,813} & 347,605\tiny{$\pm$1,934} & 598,317\tiny{$\pm$2,645} \\
NaiveRAG & \textbf{98\tiny{$\pm$4}} & \textbf{71\tiny{$\pm$3}} & \textbf{109\tiny{$\pm$5}} \\
\midrule
\name{} & \underline{316\tiny{$\pm$8}} & \underline{411\tiny{$\pm$11}} & \underline{528\tiny{$\pm$14}} \\
\bottomrule
\end{tabular}
\end{table}

In the last experiment, we computed the response time in seconds and the average context length in tokens for each benchmark. The results are reported in Table~\ref{tab:response_context}. The analysis of this table shows that the \name{}'s most notable advantage lies in its response time. In fact, \name{} achieves the fastest response time of all  methods, outperforming the fastest baseline, NaiveRAG, by 47.44\% on HotpotQA, 35.82\% on 2Wiki, and 35.14\% on MuSiQue. This is particularly significant because NaiveRAG relies on FAISS for similarity searches, and FAISS is already a highly optimized index. Outperforming this baseline confirms that dimension-aware traversal of the DAG provides a genuine computational advantage over flat index searches at full dimensionality. Compared to more complex methods, such as HippoRAG2, \name{} achieves an even greater reduction, exceeding 96.40\% across all benchmarks. 

Regarding context length, \name{} maintains moderate token usage. Although it uses slightly more tokens than NaiveRAG (7.80\% on HotpotQA and 15.78\% on 2Wiki), it is substantially more efficient than graph-based methods, such as LightRAG and GraphRAG. The disproportionately large context of these approaches derives from the KG-based retrieval pipelines, which may include extracted triples into the prompt alongside the retrieved documents. \name{} avoids this overhead entirely because it does not use a KG, reducing context size by over 79.33\% compared to GraphRAG and over 86.51\% compared to LightRAG on HotpotQA. ToG and HippoRAG2 obtain the smallest contexts by refining the evidence before generation. ToG iteratively prunes candidate entities and relations through LLM calls, and HippoRAG2 filters and re-scores the retrieved evidence through its graph-based pipeline. However, this compression does not remove the cost of selecting the evidence; rather, it shifts the cost upstream. Since Tok only counts the tokens passed to the final generative model, it does not capture the additional LLM calls and graph operations performed during retrieval, which result in response times that are more than an order of magnitude higher than those of MatRAG.

\begin{table}[ht!]
\scriptsize
\centering
\caption{Efficiency comparison across the three benchmarks in terms of response time (Res), measured in seconds, and Tok. The values represent the means and standard deviations computed over 5 samples, each composed of 1,000 different questions. For each column, the optimal value is shown in bold and the suboptimal value is underlined.}
\label{tab:response_context}
\begin{tabular}{lrrrrrr}
\toprule
 & \multicolumn{2}{c}{HotpotQA} & \multicolumn{2}{c}{2Wiki} & \multicolumn{2}{c}{MuSiQue} \\
\cmidrule(r){2-3} \cmidrule(r){4-5} \cmidrule(r){6-7}
 & Res $\downarrow$ & Tok $\downarrow$ & Res $\downarrow$ & Tok $\downarrow$ & Res $\downarrow$ & Tok $\downarrow$ \\
\midrule
HippoRAG2 & 76.84\tiny{$\pm$2.83} & \underline{591\tiny{$\pm$12}} & 80.73\tiny{$\pm$3.91} & \underline{624\tiny{$\pm$14}} & 83.12\tiny{$\pm$2.97} & \underline{671\tiny{$\pm$15}} \\
RAPTOR    & 15.23\tiny{$\pm$.41} & 854\tiny{$\pm$18} & 14.89\tiny{$\pm$.38} & 843\tiny{$\pm$17} & 15.91\tiny{$\pm$.44} & 856\tiny{$\pm$19} \\
KGP       & 31.40\tiny{$\pm$1.57} & 2,478\tiny{$\pm$31} & 26.55\tiny{$\pm$1.49} & 359\tiny{$\pm$9} & 38.04\tiny{$\pm$1.63} & 2,298\tiny{$\pm$28} \\
ToG       & 56.81\tiny{$\pm$1.74} & \textbf{124\tiny{$\pm$6}} & 40.67\tiny{$\pm$1.62} & \textbf{112\tiny{$\pm$5}} & 49.93\tiny{$\pm$1.68} & \textbf{121\tiny{$\pm$6}} \\
GraphRAG  & 15.34\tiny{$\pm$.39} & 3,812\tiny{$\pm$42} & 16.08\tiny{$\pm$.43} & 871\tiny{$\pm$19} & 16.20\tiny{$\pm$.45} & 1,398\tiny{$\pm$24} \\
LightRAG  & 20.87\tiny{$\pm$.98} & 5,841\tiny{$\pm$53} & 20.44\tiny{$\pm$.96} & 5,512\tiny{$\pm$49} & 21.03\tiny{$\pm$.91} & 7,268\tiny{$\pm$61} \\
NaiveRAG  & \underline{3.52\tiny{$\pm$.18}} & 731\tiny{$\pm$13} & \underline{3.88\tiny{$\pm$.21}} & 704\tiny{$\pm$12} & \underline{4.61\tiny{$\pm$.24}} & 814\tiny{$\pm$15} \\
\midrule
\name{} & \textbf{1.85\tiny{$\pm$.12}} & 788\tiny{$\pm$10} & \textbf{2.49\tiny{$\pm$.15}} & 815\tiny{$\pm$11} & \textbf{2.99\tiny{$\pm$.17}} & 813\tiny{$\pm$12} \\
\bottomrule
\end{tabular}
\end{table}

These results, combined with those in Tables~\ref{tab:Comparison-Results-recall} and~\ref{tab:Comparison-Results-emf1}, suggest that \name{} successfully strikes a balance between computational efficiency and contextual richness. In fact, it guarantees a considerably faster retrieval while identifying the most informative documents.

\subsection{Ablation Analysis}
\label{sub:Ablation-Analysis}

In this section, we assess the two core components of \name{} (i.e., the anchoring strategy and the Matryoshka indexing scheme) by comparing them with simpler alternatives. Unlike the hyperparameter analysis, which was performed on the validation set to avoid tuning on test data, this ablation study does not involve hyperparameter selection. Therefore, we employed the same test samples used in Section~\ref{sub:Comparison-Results} to ensure comparability with the main results. 

We started by analyzing the effect of the query anchoring strategy, controlled by $\beta_t$. Specifically, we considered  three alternatives, namely: {\em (i) } using only the original query ($q$); {\em (ii)} using only the cumulative expanded query ($q_c$); and {\em (iii)} using an adaptive combination of the two. The results are reported in Table~\ref{tab:hp-sim-mode}. This table shows that relying exclusively on either signal is suboptimal. Using only the original query ($\beta_t=1.00$) limits the ability to incorporate newly discovered contexts, thereby undermining multi-hop performance. Conversely, using only the expanded query ($\beta_t=0.00$) introduces semantic drift because irrelevant information may accumulate across iterations. The formulation proposed in \name{} (see Equation \ref{eq:beta_t}) balances these two contributions dynamically by gradually increasing the weight of the original query to maintain alignment with the initial information need. It achieves the optimal trade-off across all datasets and metrics. For instance, it increases F1 by $1.95\%$ compared to the base variant ($\beta_t=1.00)$ on HotpotQA and by $11.34\%$ on 2Wiki. It also improves F1 by $4.75\%$ over the expanded-query variant ($\beta_t=0.00$) on MuSiQue. These results highlight the importance of managing query drift in iterative retrieval and show that the proposed $\beta_t$ scheduling provides a consistent and effective scoring mechanism for different levels of reasoning complexity.

\begin{table}[!ht]
\scriptsize
\centering
\caption{Values of EM, F1, R@2 and R@5 obtained by \name{} on the three benchmarks for $\beta_t$ set to 1.00, $\frac{|{\cal R}_{t-1}|}{K}$ and 0.00. The values represent the means computed over 5 samples, each composed of $1,000$ different questions. For each column, the optimal value is shown in bold.}
\label{tab:hp-sim-mode}
\begin{tabular}{c|ccccccccc}
\toprule
 & \multicolumn{3}{c}{HotpotQA} & \multicolumn{3}{c}{2Wiki} & \multicolumn{3}{c}{MuSiQue} \\
\cmidrule(lr){2-4} \cmidrule(lr){5-7} \cmidrule(lr){8-10}
 $\beta_t$ & EM $\uparrow$ & F1 $\uparrow$ & \begin{tabular}{@{}c@{}}R@2 $\uparrow$ \\R@5 $\uparrow$ \end{tabular} & EM $\uparrow$ & F1 $\uparrow$ & \begin{tabular}{@{}c@{}}R@2 $\uparrow$ \\R@5 $\uparrow$ \end{tabular} & EM $\uparrow$ & F1 $\uparrow$ & \begin{tabular}{@{}c@{}}R@2 $\uparrow$ \\R@5 $\uparrow$ \end{tabular} \\
\midrule
1.00  & 59.22  & 72.25  & \begin{tabular}{@{}c@{}}72.10\\88.70\end{tabular} & 50.90 & 56.96  & \begin{tabular}{@{}c@{}}62.05\\76.75\end{tabular} & 25.70 & 35.92  & \begin{tabular}{@{}c@{}}41.30\\55.52\end{tabular} \\[0.5ex]
\cmidrule(lr){2-10}

$\frac{|\mathcal{R}_{t-1}|}{K}$     & \textbf{60.50}      & \textbf{73.66}  & \begin{tabular}{@{}c@{}}\textbf{72.65}\\\textbf{89.30}\end{tabular} & \textbf{52.70} & \textbf{63.42} & \begin{tabular}{@{}c@{}}\textbf{63.08}\\\textbf{79.27}\end{tabular} & \textbf{27.40} & \textbf{37.89}  & \begin{tabular}{@{}c@{}}\textbf{42.93}\\\textbf{57.08}\end{tabular} \\[0.5ex]
\cmidrule(lr){2-10}

0.00 & 58.70 & 71.24  & \begin{tabular}{@{}c@{}}72.05\\87.72\end{tabular} & 51.24 & 62.31 & \begin{tabular}{@{}c@{}}62.58\\78.81\end{tabular} & 26.42 & 36.17 & \begin{tabular}{@{}c@{}}41.17\\56.31\end{tabular} \\
\bottomrule
\end{tabular}
\end{table}

We then examined whether using shorter Matryoshka prefixes at coarser levels instead of the full embedding dimension would result in a hierarchy with lower-quality clusters. To do so, we ran HDBSCAN on all three benchmarks at each level $l$ of the hierarchy, from the coarsest clusters ($l=1$) down to the documents ($l=L$) using the truncated Matryoshka prefixes employed by \name{}. We compared the resulting clusters to those obtained by running the same procedure on full-dimensional embeddings. We used the Davies-Bouldin (DB) index~\cite{DaBo79} and the Silhouette (Sl) score~\cite{Rousseeuw87} as internal clustering quality measures. The DB index quantifies the average similarity between each cluster and its most similar clusters. Lower DB values indicate better separated and more compact clusters. The Sl score measures how close each point is to its own cluster compared to the neighboring ones. Higher Sl values denote more coherent clusters. Table~\ref{tab:clustering_quality} reports the comparison. Across all benchmarks, the two strategies produce similar clustering quality at every level. The Matryoshka prefixes even achieve slightly better DB and Sl scores in most combinations of levels and datasets. This demonstrates that truncating the representations toward coarser levels does not degrade the quality of the clusters that form the hierarchy. Therefore, the efficiency gains in response time obtained through the usage of Matryoshka prefixes and discussed in Section~\ref{sub:Comparison-Results} do not come at the expense of clustering quality.

\begin{table}[ht!]
\scriptsize
\centering
\caption{Values of the Davies-Bouldin index (DB) and the Silhouette score (Sl) for the level $l$ and embedding dimension across benchmarks. The $m_l$ column indicates the embedding dimension used for clustering with full-dimensional and Matryoshka embeddings. For each column, the optimal value is shown in bold.}
\label{tab:clustering_quality}
\begin{tabular}{ll rr rr rr}
\toprule
 & & \multicolumn{2}{c}{HotpotQA} & \multicolumn{2}{c}{2Wiki} & \multicolumn{2}{c}{MuSiQue} \\
\cmidrule(lr){3-4} \cmidrule(lr){5-6} \cmidrule(lr){7-8}
$l$ & $m_l$ & DB $\downarrow$ & Sl $\uparrow$ & DB $\downarrow$ & Sl $\uparrow$ & DB $\downarrow$ & Sl $\uparrow$ \\
\midrule
\multirow{2}{*}{1} & 768 & 0.8667 & 0.4484 & 0.7313 & 0.5512 & 0.8362 & 0.4492 \\
                   & 64  & \textbf{0.8630} & \textbf{0.4628} & \textbf{0.7198} & \textbf{0.5536} & \textbf{0.7557} & \textbf{0.5406} \\
\midrule
\multirow{2}{*}{2} & 768 & 0.8342 & 0.4663 & 0.7503 & \textbf{0.5085} & \textbf{0.7573} & 0.4746 \\
                   & 128 & \textbf{0.7616} & \textbf{0.4749} & \textbf{0.7389} & 0.5059 & 0.7661 & \textbf{0.4860} \\
\midrule
\multirow{2}{*}{3} & 768 & 0.7428 & 0.5206 & \textbf{0.6608} & 0.5390 & 0.7330 & 0.4876 \\
                   & 256 & \textbf{0.7209} & \textbf{0.5298} & 0.6890 & \textbf{0.5544} & \textbf{0.6642} & \textbf{0.5420} \\
\midrule
\multirow{2}{*}{4} & 768 & 0.6294 & 0.5601 & \textbf{0.5818} & 0.5755 & 0.6408 & 0.5561 \\
                   & 512 & \textbf{0.6131} & \textbf{0.5671} & 0.5838 & \textbf{0.5784} & \textbf{0.6315} & \textbf{0.5567} \\
\midrule
\multirow{2}{*}{5} & 768 & 0.5695 & 0.5871 & 0.5460 & 0.5849 & 0.5994 & 0.5780 \\
                   & 768 & 0.5695 & 0.5871 & 0.5460 & 0.5849 & 0.5994 & 0.5780 \\
\bottomrule
\end{tabular}
\end{table}

Finally, Tables~\ref{tab:ablation_emf1} and~\ref{tab:ablation_recall_res} report the comparison of Matryoshka's truncated and full-dimensional embeddings during retrieval. Across all available benchmarks, the Matryoshka configuration achieves better or comparable performance in terms of QA metrics and Recall while reducing response time. These results suggest that truncating embeddings to a lower-dimensional subspace, as enabled by MRL, does not negatively impact retrieval quality. This behavior can be attributed to the way Matryoshka embeddings are trained. Indeed, the most semantically relevant information is encoded in the leading dimensions, efficiently concentrating the signal and reducing noise in the higher-dimensional components. Consequently, using only the truncated prefix of the embeddings acts as a form of implicit regularization, discarding dimensions that provide little discriminative information and could introduce spurious similarity signals. We expect this effect to be particularly relevant for the clustering step. Density-based algorithms such as HDBSCAN rely on distances between points. In high-dimensional spaces these distances tend to concentrate, making density estimates and cluster separation less reliable. In MatRAG, the dimensionality decreases progressively towards the coarser levels, so clustering is performed in lower-dimensional spaces where distances remain more informative. This is consistent with Table~\ref{tab:clustering_quality}, which shows that truncated prefixes yield equal or better clustering quality in most level-dataset combinations. In particular, we obtain the largest gain at the coarsest level on MuSiQue and nearly identical scores at $l = 4$. The observed efficiency gains in response time further reinforce the practical appeal of this design choice. Operating on lower-dimensional vectors makes the retrieval step  computationally cheaper without sacrificing the quality of the retrieved documents or the downstream answers.

\begin{table}[ht]
\centering
\scriptsize
\caption{Ablation study on QA performance, as measured by EM and F1, on the three benchmarks. The \textit{Embedding} column indicates whether the system uses Matryoshka truncated (Matryoshka) or full-dimensional (Full) embeddings during retrieval. Values represent the results computed over 5 samples, composed of 1,000 different questions. For each column, the optimal value is shown in bold.}
\label{tab:ablation_emf1}
\begin{tabular}{lcccccc}
\toprule
 & \multicolumn{2}{c}{HotpotQA} & \multicolumn{2}{c}{2Wiki} & \multicolumn{2}{c}{MuSiQue} \\
\cmidrule(r){2-3} \cmidrule(r){4-5} \cmidrule(r){6-7}
Embedding  & EM $\uparrow$ & F1 $\uparrow$ & EM $\uparrow$ & F1 $\uparrow$ & EM $\uparrow$ & F1 $\uparrow$ \\
\midrule
Full  & 59.28 & 72.47 & 49.60 & 60.29 & 26.00 & 36.23 \\
Matryoshka  & \textbf{60.50} & \textbf{73.66} & \textbf{52.70} & \textbf{63.42} & \textbf{27.40} & \textbf{37.89} \\
\bottomrule
\end{tabular}
\end{table}

\begin{table}[ht]
\centering
\scriptsize
\caption{Ablation study on retrieval and efficiency performance, as measured by R@2, R@5, and Res (computed as the sum of retrieval and answer generation time) on the three benchmarks. The \textit{Embedding} column indicates whether the system uses Matryoshka truncated (Matryoshka) or full-dimensional (Full) embeddings during retrieval. Values represent the results computed over 5 samples composed of 1,000 different questions. For each column, the optimal value is shown in bold.}
\label{tab:ablation_recall_res}
\begin{tabular}{lrrrrrrrrr}
\toprule
 & \multicolumn{3}{c}{HotpotQA} & \multicolumn{3}{c}{2Wiki} & \multicolumn{3}{c}{MuSiQue} \\
\cmidrule(r){2-4} \cmidrule(r){5-7} \cmidrule(r){8-10}
Embedding & R@2 $\uparrow$& R@5 $\uparrow$& Res $\downarrow$ & R@2 $\uparrow$& R@5 $\uparrow$ & Res $\downarrow$ & R@2 $\uparrow$ & R@5 $\uparrow$ & Res $\downarrow$ \\
\midrule
Full  & 70.86 & 86.72 & 3.10 & 62.12 & 76.88 & 4.04 & 41.86 & 55.80 & 4.82 \\
Matryoshka  & \textbf{72.65} & \textbf{89.30} & \textbf{1.85} & \textbf{63.08} & \textbf{79.27} & \textbf{2.49} & \textbf{42.93} & \textbf{57.08} & \textbf{2.99} \\
\bottomrule
\end{tabular}
\end{table}

\section{Discussion}
\label{sec:Discussion}

One of the core hypotheses of \name{} is that the semantic hierarchy of a clustering structure can be exploited through the nested MRL structure. This means that coarser levels can be indexed at shorter prefix dimensions without a significant loss of discriminative power. The retrieval results support this hypothesis; in fact, \name{} outperforms the baselines on most metrics while operating with a lower dimensionality of the embeddings at the upper levels of the DAG. The largest improvements are seen in MuSiQue, which may suggest that hierarchical traversal is most beneficial when single-document similarity is insufficient to identify the relevant evidence chain. 

Another core component of \name{} is the entity-Jaccard re-ranking signal. Hyperparameter analysis (Table~\ref{tab:hp-alpha-used}) shows that the entity score complements semantic similarity. Removing it entirely ($\alpha = 1.00$) reduces performance, particularly on 2Wiki, where compositional reasoning chains tend to introduce numerous named entities across hops. Conversely, relying exclusively on the entity signal ($\alpha = 0.00$) also results in poor performance, as entity overlap alone cannot capture semantic relevance. The optimal balance is achieved at $\alpha = 0.50$ across all datasets, indicating that the two signals complement each other. Furthermore, the adaptive anchoring schedule shows that neither anchoring the search exclusively to the original query nor expanding it without control is sufficient. This is because the interpolation guided by $\beta_t$ is necessary to exploit the retrieved context and, at the same time, prevent semantic drift. This tension cannot be solved by static scoring strategies.

The clustering analysis provided further evidence in support of the central hypothesis of \name{}. As shown in Table~\ref{tab:clustering_quality}, the clusters obtained from truncated Matryoshka prefixes are competitive with those obtained from full-dimensional embeddings and tend to produce slightly better clustering quality. This is refelected in lower DB indices and higher Silhouette scores in several configurations. This suggests that, at the coarser levels of the hierarchy, the additional dimensions of the full embedding contain redundant information for forming well-separated clusters, which is removed by truncation. Consequently, the Matryoshka indexing scheme reduces the cost of similarity computations without compromising the quality of the hierarchy.

After verifying that MRL maintains the quality of the hierarchy, a complementary question is whether this also applies to end-to-end retrieval, in which truncated representations are used to traverse the DAG and rank documents. Tables~\ref{tab:ablation_emf1} and~\ref{tab:ablation_recall_res} address this issue by comparing Matryoshka's truncated and full-dimensional embeddings during retrieval. Across all benchmarks, the Matryoshka configuration achieves better performance consistently in terms of both QA metrics and retrieval recall while  reducing response time. These results suggest that truncating embeddings to a lower-dimensional subspace, as enabled by MRL, does not degrade retrieval quality. This behavior can be attributed to the way Matryoshka embeddings are trained, as the most semantically relevant information is encoded in the leading dimensions. This effectively concentrates the signal and reduces noise in the higher-dimensional components. Consequently, using only the truncated embedding prefixes acts as a form of implicit regularization, as it discards dimensions that provide little discriminative information and could introduce spurious similarity signals. The observed efficiency gains in response time further reinforce the practical appeal of this design choice, as operating on lower-dimensional vectors makes the retrieval step computationally cheaper without sacrificing the quality of the retrieved documents or the downstream answers.

The results obtained in our tests have several important implications for the design of retrieval systems. First, coupling the embedding dimension with the index depth is a productive inductive bias because it exploits the multi-scale structure of existing MRL models without requiring additional trainable parameters or training. Second, lightweight, entity-based signals can effectively proxy the reasoning and planning typically performed by LLMs, substantially reducing query time without a corresponding loss in answer quality. Third, a performance comparable to that of graph-based pipelines can be achieved with structured retrieval, without the need for explicit KGs, entity linking and LLM-based summarization. This makes \name{}'s approach applicable to a broad range of corpora without requiring explicit KG construction, eliminating the need for the preprocessing that these operations require. Together, these findings suggest that the dichotomy between cheap, flat retrieval and expensive, structured retrieval can be partially bridged by carefully co-designing the embedding strategy and the index structure.

As for the \name{}'s limitations, we observe that the entity extraction step of \name{} relies on GLiNER~\cite{Zaratiana*24}, which can fail to identify domain-specific entities or produce inaccurate extractions on corpora with unusual terminology. Errors in this step propagate into the budget control mechanism and the re-ranking signal. Additionally, the quality of the hierarchical index depends on the density of the embedding space. For instance, clusters may be poorly defined on corpora with very uniform semantic distributions, reducing the discriminative value of the DAG traversal. Finally, the greedy, top-down traversal of the DAG selects the most similar cluster centroid at each level, committing irrevocably to that branch without a backtracking mechanism. An error at a coarse level, where shorter Matryoshka prefixes provide less discriminative representations, propagates downward and may exclude relevant documents from the candidate set, regardless of their similarity at finer levels. The nearest overlapping assignment mitigates this risk by ensuring that each node is reachable from $p$ parent clusters, thus providing redundant paths through the hierarchy. However, if the correct branch is not among the selected parents of a relevant node, the traversal offers no recovery mechanism.

\section{Conclusion}
\label{sec:Conclusion}

In this paper, we have presented \name{}, a hierarchical RAG framework for multi-hop Question Answering that leverages the structural alignment between density-based clustering and MRL. \name{} organizes a document corpus into a DAG of clusters, associating each level with a correspondingly shorter MRL prefix dimension. This enables top-down retrieval that is dimensionality-aware and semantically structured. Unlike the multi-hop approach used by many KG-based RAGs, \name{} employs an entity-driven iterative mechanism that controls the retrieval budget through entity counts and re-ranks candidates based on entity overlap. Furthermore, an adaptive anchoring schedule mitigates query drift by progressively increasing the weight of the original query as the budget is consumed.

Experiments on HotpotQA, 2WikiMultiHopQA, and MuSiQue, designed to compare \name{} with seven representative baselines, demonstrate that \name{} consistently delivers superior retrieval and answer quality across a wide range of settings. Additionally, it reduces indexing time by orders of magnitude compared to graph-based methods and reduces query-time latency by over $35.14\%$ compared to NaiveRAG.

Future work will explore three main directions. The first is predicting a query-dependent retrieval budget, which adjusts the number of retrieved documents based on the query complexity. The second direction involves learning adaptive cluster connectivity, which replaces a  fixed number of parent assignments with a hierarchy that better reflects the semantic structure of the corpus. The third direction involves a query-dependent anchoring schedule that balances the original and expanded queries dynamically, enabling more effective exploration while preserving alignment with the question.

\section*{Declaration of competing interest}
The authors declare that they have no known competing financial interests or personal relationships that could have appeared to influence the work reported in this paper.

\section*{Declaration on Generative AI}
The authors declare that they used generative AI tools solely to polish the language of the manuscript, such as improving phrasing and correcting grammar.

\section*{Data availability}
The code used for our study are publicly available at the link \url{https://anonymous.4open.science/r/MatRAG}.


\bibliographystyle{plain}
\bibliography{bibliografia, temp}

\end{document}